\documentclass{article}

\usepackage{microtype}
\usepackage{graphicx}
\usepackage{subfigure}
\usepackage{booktabs} %
\usepackage{lscape}
\usepackage{amsmath}
\usepackage[ruled,vlined]{algorithm2e}

\usepackage{hyperref}

\usepackage[accepted]{icml2024}

\usepackage{amsmath}
\usepackage{amssymb}
\usepackage{amsfonts}
\usepackage{mathtools}
\usepackage{amsthm}
\usepackage{bm}

\usepackage[capitalize,noabbrev]{cleveref}

\theoremstyle{plain}

\theoremstyle{definition}

\theoremstyle{remark}

\usepackage[textsize=tiny]{todonotes}

\icmltitlerunning{Multivector Neurons: Better and Faster O(n)-Equivariant Clifford Graph Neural Networks}

\begin{document}

\twocolumn[
\icmltitle{Multivector Neurons: Better and Faster O(n)-Equivariant Clifford GNNs}

\icmlsetsymbol{equal}{*}

\begin{icmlauthorlist}
\icmlauthor{Cong Liu}{amlab,sch}
\icmlauthor{David Ruhe}{amlab,sch,an}
\icmlauthor{Patrick Forré}{amlab,sch}
\end{icmlauthorlist}

\icmlaffiliation{amlab}{AMLab, Amsterdam, Netherlands}
\icmlaffiliation{sch}{AI4Science Lab, Amsterdam, Netherlands}
\icmlaffiliation{an}{Anton Pannekoek Institute, Amsterdam, Netherlands}

\icmlcorrespondingauthor{Cong Liu}{c.liu4@uva.nl}

\icmlkeywords{Machine Learning, ICML}

\vskip 0.3in
]

\printAffiliationsAndNotice{}  %

\newcommand{\R}{\mathbb{R}}
\newcommand{\Euc}{\mathrm{E}}
\newcommand{\Ort}{\mathrm{O}}
\newcommand{\SOrt}{\mathrm{SO}}
\newcommand{\Cl}{\mathrm{Cl}}
\newcommand{\cin}{{c_\text{in}}}
\newcommand{\cout}{{c_\text{out}}}
\newcommand{\vbf}{{\mathbf{v}}}
\newcommand{\wbf}{{\mathbf{w}}}
\newcommand{\fbf}{{\mathbf{f}}}
\newcommand{\phibm}{{\bm{\phi}}}
\newcommand{\psibm}{{\bm{\psi}}}
\newcommand{\bfrak}{{\mathfrak{b}}}
\newcommand{\kbf}{{\mathbf{k}}}
\newcommand{\qbf}{{\mathbf{q}}}

\begin{abstract}
Most current deep learning models equivariant to $\Ort(n)$ or $\SOrt(n)$ either consider mostly scalar information such as distances and angles or have a very high computational complexity. In this work, we test a few novel message passing graph neural networks (GNNs) based on Clifford multivectors, structured similarly to other prevalent equivariant models in geometric deep learning. Our approach leverages efficient invariant scalar features while simultaneously performing expressive learning on multivector representations, particularly through the use of the equivariant geometric product operator. 
By integrating these elements, our methods outperform established efficient baseline models on an N-Body simulation task and protein denoising task while maintaining a high efficiency.
In particular, we push the state-of-the-art error on the N-body dataset to 0.0035 (averaged over 3 runs); an 8\% improvement over recent methods.  Our implementation is available on \href{https://github.com/congliuUvA/Multivector-Neurons}{GitHub}.
\end{abstract}

\section{Introduction}

Learning on graphs with geometric features has gathered significant attention in the field of geometric deep learning. Particularly, the integration of group equivariance to respect the symmetries in data structures yielded significant research breakthroughs. 
Among these, $\Ort(n)$- and $\SOrt(n)$-equivariance stands out as an advantageous model property when dealing with geometric data from various interdisciplinary scientific fields, such as molecular and protein structures or 3D computer vision \cite{brandstetter2022geometric, liao2023equiformer, gasteiger2022directional, deng2021vector, gao2023pifold, bekkers2024fast}.

Recently, some works \cite{brandstetter2022geometric, brandstetter2023clifford, ruhe2023geometric, ruhe2023clifford, brehmer2023geometric, liu2024clifford, zhdanov2024cliffordsteerable, spinner2024lorentz} have designed equivariant models based on the Clifford (or \emph{geometric}) algebra \cite{ruhe2023clifford}: Clifford Group Equivariant Neural Networks (CGENNs). 
These models embed geometric features in the Clifford algebra vector space and, based on linear combinations of \emph{multivectors} (elements of the Clifford algebra) and geometric products, they achieve $\Ort(n)$- or $\SOrt(n)$-equivariance. 
In a sense, CGENNs are similar to standard feedforward neural network but each neuron is multivector-valued.
Hence: \emph{Multivector Neurons}.
On several benchmarks, CGENNs (in particular, their graph variants) have achieved excellent results.
However, their heavily parameterized geometric product layers are computationally intense and consume significant memory. 

In this work, we test a novel set of fast and expressive Clifford graph neural networks (GNNs), incorporating parameterizations akin to those used by Equivariant Graph Neural Networks (EGNN) \cite{satorras2022en}, Geometric Vector Perceptron (GVP) \cite{jing2021learning}, and Vector Neurons \cite{deng2021vector}. 
In our design choices, instead of relying on expensive geometric product layers in \cite{ruhe2023clifford} for expressiveness, we incorporate more cheap and efficient multi-layer perceptrons operating on invariant scalar features, while ensuring group equivariance through the representation of geometric features as multivectors. 
Consequently, our approach yields a suite of networks that are both expressive and efficient, making them suitable for large-scale geometric graph applications. 

We demonstrate the efficacy of these models in two settings: an N-body simulation task (achieving new state of the art predictive results) and a protein denoising task.
The promising results on the protein denoising task makes these models suitable for denoising diffusion generative modeling, which is a promising avenue for future work.

\section{Clifford Group Equivariant Neural Networks}
CGENNs are based on a powerful mathematical tool: The \textit{Clifford} or \textit{geometric} algebra. The Clifford algebra $\Cl(V, \mathfrak{q})$ is generated by a vector space $V$ coupled with a quadratic form $q: V \rightarrow \mathbb{R}$. 
In this work, we restrict $V:= \R^3$ and $\mathfrak{b}$ to be the standard (Euclidean) inner product $b(v) := \langle v, v \rangle$.
As such, we will write $\Cl(\R^3)$.
The Clifford algebra has associated a product, the \emph{geometric product}, which can multiply vectors to generate higher-order objects like bivectors.
\emph{Multivectors} $\vbf \in \Cl(\R^3)$ are general Clifford elements defined as $\vbf = \sum_{i \in I}c_iv_{i,1}* \dots *v_{i,k}$, where $I$ is finite, $v_{i,j} \in \mathbb{R}^d$ and $c_i \in \mathbb{R}$. 
I.e., they are finite linear combinations of geometric products of vectors, denoted with $*$.
The Clifford algebra has a bilinear form extended from the usual one $\mathfrak{b}: \Cl(\R^3) \times \Cl(\R^3) \to \R$ and quadratic form $\mathfrak{q}(\vbf) := \mathfrak{b}(\vbf,\vbf)$.

The Clifford algebra can be decomposed into orthogonal subspaces. 
Among these spaces, we have the scalars $\R$ and also the original vector space $\R^3$.
As such, \citet{ruhe2023clifford} embed vector- or scalar-valued data in the algebra.
Then, the data gets processed as multivectors through linear combinations and geometric products using carefully designed neural network layers.
In general, these layers are $\Ort(n)$-equivariant.
In this work, we make a sharper distinction between scalars and multivector-valued data, processing the scalars for fast and expressive representation learning, while updating multivectors to maintain the ability to build sophisticated geometric representations.
We do so by taking inspiration from established equivariant architectures.

\section{Methods}

\subsection{Clifford-EGNN}
Our first tested architecture extends EGNN \cite{satorras2022en}.
We consider a graph where each node $i$ is associated with a position $x_i \in \R^3$ and node feature $h_i \in \mathbb{R}^c$.
EGNNs take $l$-th layer node embeddings $h_i^l$ and node positions $x_i^l \in \mathbb{R}^3$ to output $l+1$-th layer node embeddings and positions.
The message passing scheme is as follows:
\begin{align}
m_{ij}^l &= \phi_e(h_i^l, h_j^l, ||x_i^l - x_j^l||) \\
m_i^l &=  \sum_{j \in \mathcal{N}_i} m_{ij}^l \\
x_i^{l+1} &= x_i^l + \sum_{j \in \mathcal{N}_i} \phi_x(m_{ij}^l) \cdot (x_i^l - x_j^l)  \\
h_i^{l+1} &= \phi_h(h_i^l, m_i^l)
\end{align}
with $\phi_e, \phi_x, \phi_h$ as multi-layer perceptrons acting as edge message networks, position message networks and node features update networks, where $\mathcal{N}_i$ represents the set of neighbor nodes to node $i$.

In the Clifford counterpart (details decribed below), we use invariant messages to scale equivariant features while processing arrays of multivectors in a feedforward manner. We have several multivector channels that interchange information with a scalar network. In update steps, we apply equivariant geometric product layers on multivectors to refine the multivector features and exchange geometric information multiplicatively.

After an initial linear embedding layer, we can consider an array of multivectors $\vbf \in \left(\Cl(\R^3)\right)^c$, where $c$ denotes the number of channels.
Furthermore, we have scalar features $h \in \mathbb{R}^s$.
The message passing scheme from node $j$ to node $i$ is defined as follows.
\begin{align}
\vbf_{ij}^l &= \phibm_e^l(\vbf_i^l- \vbf_j^l) \\ \label{eq1}
s_{ij}^l &= \left[h_i^l, h_j^l, \mathfrak{q}(\vbf_{ij}^l)\right] \\
m_{ij}^l &= \phi_e^l(s_{ij}^l) \\
m_i^l &=  \sum_{j \in \mathcal{N}_i} m_{ij}^l \\
\vbf^{l+1} &= \vbf_i^l + \frac{1}{\sqrt{|\mathcal{N}_i|}}\psibm_v^l\left(\sum_{j \in \mathcal{N}_i} \phi_v^l(m_{ij}^l) \cdot \vbf_{ij}^l \right)\\
h_i^{l+1} &= \phi_h^l(h_i^l, m_i^l) \label{eq2}
\end{align}
where $\left[\cdot, \dots, \cdot, \right]$ denotes concatenation.
Note that we consider the positions simply as vector-valued features that we embed in a multivector, together with other vectors if available.
Here, $\phibm$ and $\psibm$ both denote multivector-valued layers, where the latter one in addition to linear layers also applies (unparameterized) geometric products \cite{ruhe2023clifford} that create higher-order elements in Clifford subspaces, providing the ability to capture complex geometric features. 
$\phi$ denote regular MLP layers operating in scalars.
These invariant networks are cornerstones of the expressive power and scalability of the Clifford-EGNN.

\subsection{Multivector Neurons and MVN-GNN}
Vector neurons (VN) typically operate on point cloud data $x \in \left(\mathbb{R}^n\right)^\cin$, but can be extended to arbitrary vectors $v$.
The core idea is to linearly combine channels of vectors, and apply a vector nonlinearity, completely analogous to regular feed-forward architectures.
That is, 
\begin{align}
    v'_i = L(v; W)_i = \sum_{j=1}^\cin w_{ij} v_j
\end{align}
parameterized by weight matrix $W \in \mathbb{R}^{\cout \times \cin}$, where $w_{ij}$ is the element in $i$-th row, $j$-th column of weight matrix $W$.

To maintain the expressiveness of the networks other than using only linear combinations of vectors, VN defines an $\Ort(n)$-equivariant activation function. 
Given a vector feature  $v \in ({\mathbb{R}^n})^c$, weight matrix $Q \in \mathbb{R}^{c \times c}$, weight matrix $K \in \mathbb{R}^{c \times c}$, the output $v'$ of activation function is defined as:
\begin{align}
    q &= L(v; Q), & k &= L(v; K),
\end{align}
\begin{equation}
    v_i' = 
    \begin{cases} 
        q_i & \text{if } \langle q_i, k_i \rangle \geq 0 \\ 
        q_i - \left\langle q_i, \frac{k_i}{\|k_i\|} \right\rangle \frac{k_i}{\|k_i\|} & \text{otherwise\,,}
    \end{cases}
\end{equation}
where $\left\langle \cdot,\cdot \right\rangle$ defines the inner product in Euclidean space. 
The second case is alwo known as the \emph{rejection} of $q$ from $k$.
Note that there are no scalar inputs in vector neurons, and this hinders the usage of vector neurons in learning on geometric graphs with scalar features. 

In Multivector Neurons (MVN), we use similar ideas but instead operate on multivectors. 
Essentially, we apply linear layers analogous to the ones from Vector Neurons and replace the vector inner product $\langle \cdot, \cdot \rangle$ with its Clifford algebra-extended version $\mathfrak{b}$.
That is,
\begin{equation}
    \vbf_i' = 
    \begin{cases} 
        \qbf_i & \text{if } \bfrak \left( \qbf_i, \kbf_i \right) \geq 0 \\ 
        \qbf_i - \mathfrak{b} \left(\qbf_i, \frac{\kbf_i}{\sqrt{\bfrak(\kbf, \kbf)}}\right) \frac{\kbf_i}{\sqrt{\bfrak(\kbf, \kbf)}} & \text{otherwise\,.}
    \end{cases}
    \label{eq:mvn_nonlin}
\end{equation}
Note that this nonlinearity can be extended to more generally use the \emph{rejection} of multivectors \cite{chisolm2012geometric}.

As such, a simple two-layer Multivector Neuron MLP (MVN-MLP) then has
\begin{align}
    \vbf' = \phibm_2 \cdot \bm{\sigma}_1 \cdot \phibm_1(\vbf)\,,
\end{align}
where $\bm{\sigma}$ denotes the nonlinearity as introduced in \Cref{eq:mvn_nonlin} and $\phibm$ denotes a multivector linear layer.
Additionally, similar to Clifford EGNN, we also introduce scalar networks such that MVN can accept invariant inputs. 
Formally, with central node index $i$, neighbor node index $j$, invariant scalar features $h \in \mathbb{R}^{s}$, multivector features $\vbf$, invariant scalar message networks $\phi_e$, invariant vector message networks $\phi_v$, MVN-MLP, quadratic form $\mathfrak{q}$ in Clifford space, invariant scalar update networks $\phi_h$, geometric product layers $\psibm_v$, layer index $l$, we can define the message passing scheme in MVN as follows:
\begin{align}
\vbf_{ij}^l &= \text{MVN-MLP}^l(\left[\vbf_i^l, \vbf_j^l\right]) \\ 
s_{ij}^l &= \left[h_i^l, h_j^l, \mathfrak{q}(\vbf_{ij}^l)\right] \\
m_{ij}^l &= \phi_e^l(s_{ij}^l) \\
m_i^l &=  \sum_{j \in \mathcal{N}_i} m_{ij}^l \\
\vbf_i^{l+1} &= \vbf_i^l + \frac{1}{\sqrt{|\mathcal{N}_i|}}{\psibm_v^l}\left(\sum_{j \in \mathcal{N}_i} \phi_v^l(m_{ij}^l) \cdot \vbf_{ij}^l\right) \\
h_i^{l+1} &= \phi_h^l(h_i^l, m_i^l)
\end{align}
Again, we treat node positions $x_i$ as vector features, which we process through their multivector embedding.

\subsection{Multivector Perceptrons and MVP-GNN}
We again assume a graph where we have scalar node features $h \in \mathbb{R}^{s}$ but also vector features $v \in (\mathbb{R}^3)^c$.
Geometric vector perceptrons \cite{jing2021learning} (GVP) define a layer that operates on both scalars (invariants) and vectors.
Invariant inputs $h \in \mathbb{R}^{s}$ are processed by MLPs $\phi$.
Vector features $v \in (\mathbb{R}^3)^c$ are processed by weight matrix $W \in \mathbb{R}^c \times \mathbb{R}^c$ acting as a linear combinator of the input vector features. 
GVP establishes communication between vectors and scalars by using norms of vectors as invariant input to scalar networks.
With invariant scalar feature $s$, vector features $v$, invariant scalar network $\phi$, vector weight matrices $W_h, W_\mu$, sigmoid function $\sigma$, a GVP layer is defined as follows:
\begin{align}
(v_\mu, v_h) &= L(v, W_\mu), \, L(v, W_h) \\
(s_\mu, s_h) &= ||v_\mu||, \, ||v_h|| \\ \label{norm}
(s', v') &= \phi_s(s_h, s),\,  \sigma(s_\mu) \cdot v_\mu 
\end{align}
Then, for a GVP-style message passing layer, one can use GVPs as building blocks for message and update functions.
Similar as EGNN, GVP applies MLPs to invariant scalars, scaling and linear combinations to vectors, thus the output scalars are invariant to transformations and output vectors are equivariant to $\Ort(n)$.

\begin{table*}[]
    \centering
    
\begin{tabular}{lccc}
    \toprule
    Models & MSE ($\downarrow$) & Seconds/Iteration ($\downarrow$) & Memory (MiB) ($\downarrow$) \\ \midrule
    EGNN \citep{satorras2022en}   & $0.0076 \pm 0.0003$  &      $0.0115 \pm 0.0000$  & $334$   \\ 
    GVP-GNN \citep{jing2021learning} & $0.0069 \pm 0.0002$  &    $0.0152 \pm 0.0001 $ & $370$     \\ 
    CGENN \citep{ruhe2023clifford}     & $0.0039 \pm 0.0000$   &  $0.3240 \pm  0.0000$ & $814$    \\ 
    WeLNet \citep{hordan2024} & $0.0036 \pm 0.0002$ & $0.0183 \pm 0.0001$ & $1322$ \\ \midrule
    Clifford EGNN   & $0.0038 \pm 0.0001$     &   $0.0211 \pm 0.0001$    & $336$   \\
    MVP-GNN   & $\mathbf{0.0035} \pm 0.0000$    &       $0.0431 \pm 0.0000$ & $522$  \\ 
    MVN-GNN  & $0.0036 \pm 0.0001$     &    $0.0274 \pm 0.0000$  & $430$   \\ \bottomrule
\end{tabular}

    \caption{MSE ($\downarrow$) of the tested models on the N-body simulation dataset.}
    \label{tab:nbody}
\end{table*}

\begin{table*}[]
    \centering
    \begin{tabular}{lccc}
    \toprule
    Models & MSE ($\downarrow$) & Seconds/Iteration ($\downarrow$) & Memory (MiB) ($\downarrow$) \\ \midrule
    EGNN \citep{satorras2022en}   & $0.392$  &      $0.018$  & $6558$   \\ 
    GVP-GNN \citep{jing2021learning} & $0.678$  &    $0.036 $ & $11966$     \\ 
    CGENN \citep{ruhe2023clifford}     & N/A   &  N/A     & OOM \\  \midrule
    Clifford EGNN   & $0.244$     &   $0.045$    & $8508$   \\
    MVP-GNN   & $\mathbf{0.239}$    &       $0.178$ & $24800$  \\ 
    MVN-GNN  & $\mathbf{0.239}$     &    $0.109$  & $20530$   \\ \bottomrule
\end{tabular}

    \caption{MSE ($\downarrow$) of the tested models on the protein backbone structure denoising}
    \label{tab:protein}
\end{table*}
Correspondingly, in Linear Multivector Perceptrons (MVP-Lin), we have an multivector $\vbf$, multivector feedforward layers $\phibm_\mu, \phibm_h$, scalar networks $\phi$ with a feedforward layer defined as follows:
\begin{align}
(\vbf_\mu, \vbf_h) &= \phibm_\mu(v), \phibm_h(v) \\ 
(s_\mu, s_h) &= \sqrt{\mathfrak{q}(\vbf_\mu)}, \sqrt{\mathfrak{q}(\vbf_h)} \\
(s', \vbf') &= \phi(s_h, s),\,  \sigma(s_\mu) \cdot \vbf_\mu 
\end{align}
To create higher-order subspace elements, with  
 geometric product network $\psibm$, multivector feedforward layers $\phibm$, scalar networks $\phi$, we also define geometric product Multivector Perceptrons (MVP-GP) as follows:
\begin{align}
    \wbf &= \psibm(\vbf)  \\
    \vbf' &= \phibm(\wbf + \vbf) \\
    s' &= \phi\left(\left[s, \vbf'^{(0)} \right]\right), \,\vbf'^{(0)} = s'
\end{align}
As such, an MVP-style message passing layer can be defined as:
\begin{align}
\vbf_{ij}^l &= \phibm_e\left([\vbf_i^l, \vbf_j^l]\right)  \\
s_{ij}^l &= \phi_e\left([s_i^l, s_j^l]\right) \\
(s_{ij}^{l'}, \vbf_{ij}^{l'}) &= (\text{MVP-GP}_e \cdot \text{MVP-Lin}_e)\left(s_{ij}^l, \vbf_{ij}^l\right) \\
(s_i^{l'}, \vbf_i^{l'}) &= \frac{1}{\sqrt{|\mathcal{N}_i|}} \sum_{\mathcal{N}_i} s_{ij}^{l'},  \frac{1}{\sqrt{|\mathcal{N}_i|}}\sum_{\mathcal{N}_i} 
\vbf_{ij}^{l'} \\
(s_i^{l+1}, \vbf_i^{l+1}) &= (\text{MVP-GP}_v \cdot \text{MVP-Lin}_v)\left(\left[s_i^l, s_i^{l'}\right], \left[\vbf_i^l, \vbf_i^{l'}\right]\right) 
\end{align}

\section{Experiments}
We apply our models, Clifford EGNNs, MVN-GNN, and MVP-GNN, on two benchmark datasets: the N-Body simulation dataset \cite{satorras2022en} and SideChainNet \cite{king2020sidechainnet}.

\subsection{N-Body System}
The N-Body simulation dataset is a popular benchmark for evaluating equivariant deep learning models. The task is to use the initial positions, charges, and velocities of $n = 5$ charged particles to predict their final positions after 1000 time steps. Following \citet{satorras2022en}, we simulate 3000 samples for the training set, 2000 samples for the validation set, and 2000 samples for the test set. 
We use 100 graphs per batch, a learning rate of $5 \times 10^{-3}$, and a weight decay of $1 \times 10^{-4}$ to train our Clifford models.
The Mean Square Error (MSE) from our Clifford models and baseline models are presented in \Cref{tab:nbody}. Our Clifford models outperform all baseline models in terms of MSE. 
Although our tested Clifford models consume more memory and computational time, they are much more efficient than the Clifford Group Equivariant Neural Networks (CGENN) from \citet{ruhe2023clifford}. 
We push the state-of-the-art error on the N-body dataset to 0.0035 (averaged over 3 runs); an 8\% improvement over recent methods.
These experiments are conducted on an NVIDIA RTX A100 GPU.

\subsection{Protein Structure Denoising}
SideChainNet \cite{king2020sidechainnet} is a dataset containing all atoms in protein structures for machine learning research.
For our experiment, the task is to recover noised 3D protein structures by predicting the positions of each atom. 
This is inspired by the GitHub user \textsc{lucidrains} who benchmarks several equivariant architectures using this task \cite{wang2024github}.
For simplicity, we choose to work only with backbone atoms C, N, and O. 
For a set of backbone atoms $\{C_i, N_i, O_i\}_{i=0}^n$, we add Gaussian noise to the coordinates of the backbone atoms. 
We use the noised backbone atom coordinates as input and use equivariant models to predict the original backbone atom coordinates. 
For each graph, we connect every node with $k=16$ nearest neighbors. 
We use 16 graphs per batch with a learning rate of $1 \times 10^{-3}$ to train our Clifford models. 
We choose MSE as our metric to evaluate the models' performance. From \Cref{tab:protein}, we observe that Clifford models surpass the baselines by a large margin. 
Clifford EGNN, in particular, uses even less memory than GVP-GNN for training. These results show that our Clifford models better grasp geometric features while being fast and memory efficient compared to CGENN. These experiments are conducted on an NVIDIA RTX A6000 Ada GPU.

\section{Conclusion}
In this work, we tested a set of fast and expressive Clifford models: Clifford EGNN, MVP-GNN, and MVN-GNN, based on the design of EGNN, GVP, and VN. 
These are all equivariant to $\mathrm{O}(n)$.
These new parameterizations rely more heavily on fast and expressive scalar networks, that occasionally interact with multivectors that contain higher-order geometric information.
We observe promising or even state of the art results on N-Body simulation tasks and protein structure denosing task, while sometimes being more efficient than the baseline models. 
Furthermore, this work provides potential directions for the design of Clifford networks that are suitable for large scale graph learning tasks.

\section*{Acknowledgments}
We would like to thank Maksim Zhdanov for all the helpful discussions.

\bibliography{example_paper}
\bibliographystyle{icml2024}
\newpage
\appendix
\onecolumn

\end{document}